\documentclass[runningheads]{llncs}
\usepackage[table]{xcolor}
\usepackage{graphicx}
\usepackage{subfigure}
\usepackage{booktabs}
\usepackage{url}
\usepackage{amsmath,amssymb,amsfonts}
\usepackage{multirow}
\usepackage{wrapfig}
\usepackage{xcolor}
\usepackage{bbding}
\definecolor{digitization}{RGB}{66,118,170}
\definecolor{blur}{RGB}{23,138,161}
\definecolor{color}{RGB}{119,132,149}
\definecolor{stain}{RGB}{94,92,162}
\begin{document}
	\title{Benchmarking the Robustness of Deep Neural Networks to Common  Corruptions in Digital Pathology}
	\newcommand{\repeatthanks}{\textsuperscript{\thefootnote}}
	\author{
		Yunlong Zhang\inst{1,2}\thanks{equal contribution} \and
		Yuxuan Sun\inst{1,2}\repeatthanks \and
		Honglin Li\inst{1,2} \and
		Sunyi Zheng\inst{2} \and
		Chenglu Zhu\inst{2} \and
		Lin Yang\inst{2(}\Envelope\inst{)}
	}
	
	\institute{College of Computer Science and Technology, Zhejiang University \and
		School of Engineering, Westlake University \\
		yanglin@westlake.edu.cn
	}
	\maketitle              
	\begin{abstract}
		
		When designing a diagnostic model for a clinical application, it is crucial to	
		guarantee the robustness of the model with respect to a wide range of image corruptions. Herein, an easy-to-use benchmark is established to evaluate how deep neural networks perform on corrupted pathology images. Specifically, corrupted images are generated by injecting nine types of common corruptions into validation images. Besides,  two classification and one ranking metrics are designed to evaluate the prediction and confidence performance under corruption. Evaluated on two resulting benchmark datasets, we find that (1) a variety of deep neural network models suffer from a significant accuracy decrease (double the error on clean images) and the unreliable confidence estimation on corrupted images; (2) A low correlation between the validation and test errors while replacing the validation set with our benchmark can increase the correlation. Our codes are available on \url{https://github.com/superjamessyx/robustness_benchmark}.

		\keywords{Robustness  \and Digital pathology \and Benchmark \and Corruption.}
	\end{abstract}
	\section{Introduction}
	Deep neural networks (DNNs) have recently made significant advances to a variety of computer vision tasks \cite{dosovitskiy2020image,he2016deep,ronneberger2015u}. Nevertheless, 
	it has been revealed that DNNs are vulnerable to input corruptions \cite{geirhos2018imagenet,hendrycks2019benchmarking,szegedy2013intriguing,azulay2018deep,dodge2016understanding}. In digital pathology image analysis, DNNs will be further affected by corruptions compared to natural image processing due to the following reasons: (1) The complex imaging processes, including tissue processing, cutting, staining, scanning, and storage \cite{barisoni2020digital}, often generate more severe corruptions; (2) Interclass differences in pathology images are smaller and blurrier than those in natural images, causing the decision boundary relies more on details and is more susceptible to corruptions. Moreover, diagnostic systems must be robust to corruption since their predictions influence or even determine subsequent treatment decisions. Therefore, studying a robust model against image corruption is crucial for pathology image analysis.
	
	It is pivotal to design a model evaluation criterion prior to building robust models. However, it is impossible to test models against all possible corruption types. To remedy this, we propose to evaluate models on certain types of corruptions that commonly occur in pathology images. More specifically, a new benchmark is presented to evaluate the model performance on nine types of common corruptions, each spanning five levels of severity. These corruptions are easy-to-use in practical settings since they are implemented by easy image processing techniques. We apply the proposed corruptions to the validation set of two large multi-center datasets, generating two benchmark datasets, Patchcamelyon-C and LocalTCT-C. Moreover, two classification and one ranking metrics are designed to evaluate the stability of the prediction (i.e., the accuracy of classification) and of the confidence (i.e., probabilities associated with the predicted class). 
	
	With the proposed benchmark, we evaluate the performance of ten convolutional neural networks (CNNs) and three vision transformers. Experimental results indicate that (1) the error rate on corrupted images is approximately doubled on clean images. Even though DNNs have been constantly improved over the past decade, their classification performance on corrupted pathology images changes slightly; (2) All models show unreliable confidence estimations while the robustness of the confidence seems to be slightly opposite to that of the prediction; (3) When applying models to practical scenarios, the validation error has a low correlation with the test error (Pearson’s Correlation r = -0.02 on Patchcamelyon), failing in providing an unbiased evaluation of a model in general machine learning algorithms. The error on our benchmark is more correlated with the test error (improving Pearson’s Correlation on Patchcamelyon to 0.45), illustrating our benchmark is better suited than the validation set in evaluating the generalization ability of the model.
	
	\noindent\textbf{Related work.} Several studies have revealed that DNNs are vulnerable to common corruptions, such as blur and Gaussian noise \cite{dodge2016understanding} and translations \cite{azulay2018deep}. In the field of digital pathology, few studies have also confirmed that the performance of DNNs drops a lot when encountering bubbles, uneven illumination, and so on \cite{wang2021stress}. To systematically study and improve the robustness of DNNs against corruptions, corruption benchmarks were first presented in the field of image recognition \cite{hendrycks2019benchmarking} and were further extended to object detection \cite{michaelis2019benchmarking}, semantic segmentation \cite{kamann2021benchmarking}, pose estimation \cite{wang2021human}, and person Re-ID \cite{chen2021benchmarks}. Our benchmark is inspired by them but is specially designed for pathology images in the aspects of the corruption type and metric choices.

	The confidence plays an important role in clinical diagnosis. For instance,  when confidence in a network for disease diagnosis is poor, control should be transferred to doctors \cite{jiang2012calibrating}. However, the unreliable confidence becomes the critical weakness of most DNNs \cite{guo2017calibration}. In our benchmark, a new metric to evaluate the stability of confidence under corruption is proposed.

	\section{Benchmark Design}
	\subsection{Formulation}
	
	\noindent\textbf{Prediction error under corruptions.} Assume we have a classifier $f$ trained on samples from a distribution $\mathcal{D}$. Most existing studies supposed test samples drawn from the same distribution, i.e., $\{(x, y)\}\sim \mathcal{D}$, and evaluated their methods by calculating $\mathbb{P}_{(x;y)}(f(x) = y)$. Yet in a vast range of cases the classifier is tasked with classifying low-quality or corrupted
	inputs.  In this view, a more suitable way to evaluate the error of the classifier is
	\begin{equation}\label{eq1}
		\mathbb{E}_{c\in C}[\mathbb{P}_{(x;y)}(f(c(x)) = y)].
	\end{equation} 
	where $C$ denotes a set of corruption functions that will occur in pathology images.

	\noindent\textbf{Confidence ranking under corruptions.}  Intuitively, corruptions will undermine the confidence, causing samples with more severe corruptions are predicted with less confidence. We would like the confidence estimation of DNNs, i.e., $\hat p(x) = \max \mathbb{P}(f(x)|x)$, to match human intuition. Thus,  confidence values should obey 
	\begin{equation}\label{eq2}
		\hat p(c(x, s1)) > \hat p(c(x, s2)), s.t. 0 \leq s1 < s2 \leq 5,
	\end{equation} 
	where $c(x, s1)$ and $c(x, s2)$ denote corruption samples with severity levels $s1$ and $s2$, respectively. Besides, the sample with the severity level of 0 denotes the clean one, i.e., $c(x, 0)=x$.

	The corruption functions $C$ is the core in Eq. \ref{eq1} and Eq. \ref{eq2}. However, defining all possible $C$ is unfeasible because the computational cost grows proportionally to the number of corruption functions $|C|$. Hence, we design the corruption functions that often occur in pathology images in Sec. \ref{sec22}. Moreover, in Sec. \ref{sec23}, two classification and one ranking metrics are proposed to measure the stability of the prediction and of the confidence.
	
	\begin{figure*}[!h]
		\centering
		\includegraphics[width=\columnwidth]{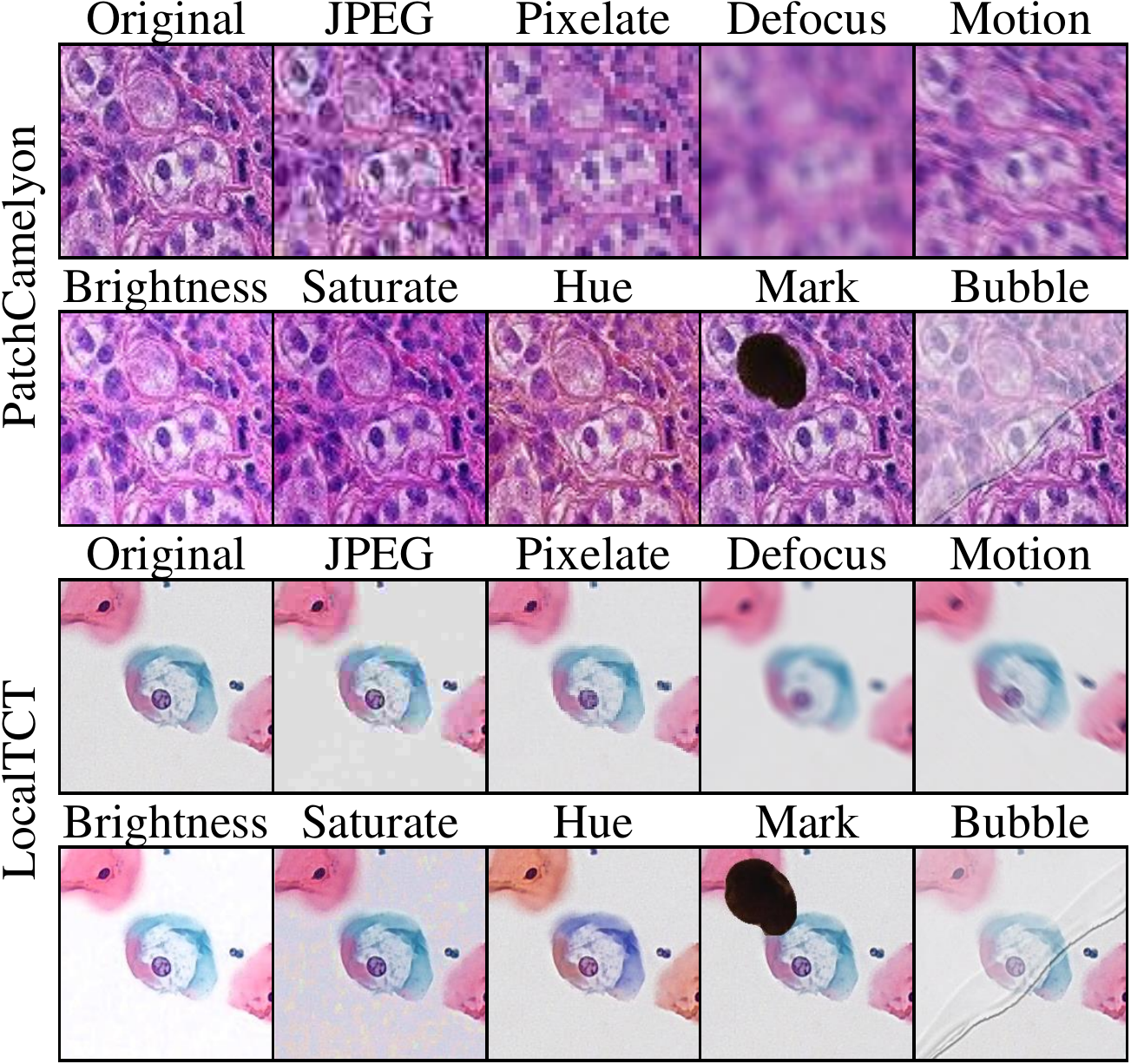}
		\caption{Two examples from PatchCamelyon and LocalTCT are imposed to nine types of corruptions that commonly occur in pathology images.}
		\label{fig1}
	\end{figure*}

	\subsection{Corruption setup} \label{sec22}
	The presented corruptions consist of several types of Digitization: \textit{JPEG} and \textit{pixilation}; Blur: \textit{defocus blur} and \textit{motion blur};
	Color: \textit{brightness}, \textit{saturation}, and \textit{hue}; and Stain: \textit{mark} and \textit{bubble} (illustrated in Fig. \ref{fig1}).These types of corruptions are then introduced in detail.
	
	The \textit{JPEG} is a lossy image compression format and imitates the image distortion brought by different storage formats. The common image compression methods in digital slides include JPEG, JPEG2000, JPEG-XR, H.265, and JPEG-LS \cite{liu2017current}. The \textit{pixilation} occurs when images are enlarged beyond the true resolution \cite{rohde2014carnegie}. As an example, when the classifier is trained on the data at $40\times$ magnification and tested on the data at $20\times$ magnification, the \textit{pixilation} will occur when test images are zoomed in twice  in the evaluation process. The \textit{defocus blur} occurs in the situation of insufficient focusing accuracy of the auto-focusing systems \cite{farahani2015whole}. Similarly, the \textit{motion blur} results from sample movement during camera exposure \cite{farahani2015whole}. The \textit{brightness}, \textit{saturation}, and \textit{hue} vary in the cases of different illuminations, scanners, stain concentration or even time elapse \cite{clarke2017colour}, severely degrading the generalization ability of DNNs \cite{yamashita2021learning}. The \textit{mark} and \textit{bubble} are two typical stains in pathology images. The \textit{mark} appears when pathologists use marking pens to delineate regions of interest on a pathology slide, for measurement, or other uses \cite{wang2021stress}. The \textit{bubble} is introduced by air getting underneath the coverslip \cite{taqi2018review}.
	
	\begin{figure*}[!h]
		\centering
		\includegraphics[width=\columnwidth]{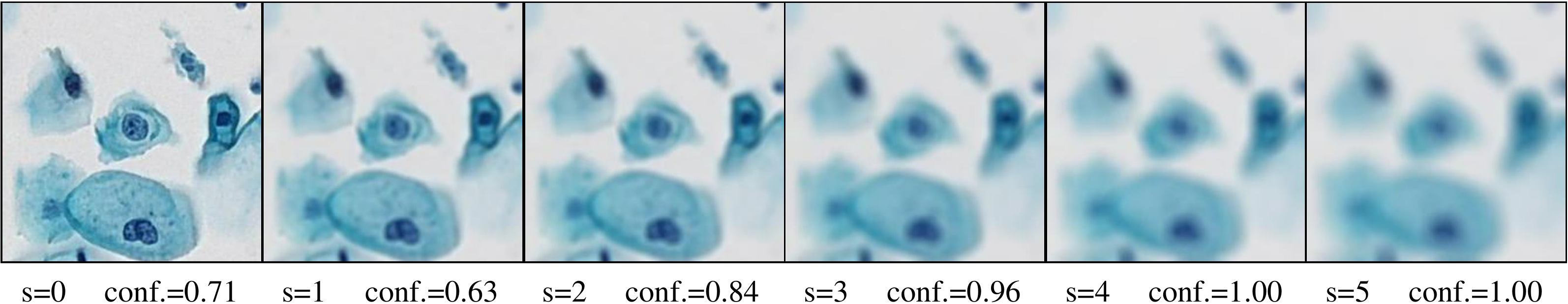}
		\caption{The severity level of corruption rises from left to right, causing the cell morphology becomes blurred gradually. Pathologists will reduce the confidence in diagnosing it as LSIL, which does not match the predictive confidence values in bottom right corner. Better viewed with zooming in.}
		\label{fig2}
	\end{figure*}
	
	Moreover, each type of corruption has five levels of severity to manifest themselves at varying intensities. One example with five different severity levels of the \textit{defocus blur} is given in Fig. \ref{fig2}. All types of corruptions are implemented by functions so that they can be plugged into the dataloader for portability and storage saving. Although these corruptions are not equivalent to real-world ones, model performance on them can also reflect the adaptive ability of models to corruption. In total, 45 corruptions are applied to validation data for evaluating the corruption robustness of a pre-existing network.

	\subsection{Metric setup} \label{sec23}
	\textbf{Corruption Error (CE).} Eq. \ref{eq1} can be realized by aggregating errors on all types and severity levels of corruptions in corruption functions $C$ defined in Sec \ref{sec22}. The corruption error of the prediction is abbreviated by CE and formulated as follow: 
	\begin{equation}\label{eq4}
		\text{CE} = \frac{1}{N_c * N_s}\sum_{c=1}^{N_c}\sum_{s=1}^{N_s} E^{c,s},
	\end{equation} 
	where $E^{c,s}$ denotes the error on corruption type $c$ at level of severity $s$, while $N_c=9$  and $N_s = 5$ indicate the number of corruptions and severity levels, respectively.

	\noindent\textbf{Relative Corruption Error (rCE).} We also focus on the degraded performance resulting from corruptions. To measure the relative degradation of performance on corrupted data compared to clean data, the rCE is defined as
	\begin{equation}\label{eq5}
		\text{rCE} = \frac{\text{CE}}{\text{Error}},
	\end{equation} 
	where Error is the error on the validation set without corruption. Investigating which components are relative to the rCE is essential for further improving model robustness against corruptions.

	\noindent\textbf{Corruption Error of Confidence (CEC).} Predictive confidence values may not match pathologist's intuition that confidence values fall with corruption severity levels increasing (e.g., the example in Fig. \ref{fig2}). To  measure this mismatch, a new metric, CEC, based on the  number of swap operations in the bubble sort method is presented. Specifically, with the confidence sequence $S(x)$  denoted by $\{\hat p(c(x, 0)),\dots,\hat p(c(x, N_s))\}$, the CEC is defined as 
	\begin{equation}\label{eq6}
		\text{CEC} = \frac{1}{N_v * N_c}\sum_{c=1}^{N_c}\sum_{x\in \mathcal{V}}\frac{K_{\tau}(S(x))}{C_{N_s}^2},
	\end{equation} 
	where $\mathcal{V}$ denotes the validation set, Kendall Tau $K_{\tau}()$ \cite{fagin2003comparing} denotes the number of swap operations when sorting the confidence sequence from largest to smallest with the bubble sort method and $C_{N_s}^2$ is the comparison times between any two confidence values in $S$.

	\section{Experiments}
	\subsection{Experimental setup}
	\textbf{Corruption implementation.} Here, implementations of corruptions are introduced in brief. The \textit{JPEG} saves the image to jpeg format and then reloads it. The \textit{pixilation} zooms in the image and then zooms out it to its original shape. The \textit{defocus blur} and \textit{motion blur} filter the image with two customized kernels. The \textit{brightness}, \textit{saturation}, and \textit{hue} modify the image in the corresponding channel of the HSV space. The \textit{mark} and \textit{bubble} mix the image with predefined matrix of mark and bubble examples, respectively.
	
	\noindent\textbf{Datasets.}
	To evaluate the robustness against corruptions, one histopathological dataset, i.e., PatchCamelyon\cite{veeling2018rotation}, and one cytological dataset, i.e., LocalTCT, are adopted. The PatchCamelyon is derived from the Camelyon 16 challenge\cite{bejnordi2017diagnostic}. It is divided into a training set of 262,144 examples, and a validation and test set both of 32,768 examples. Each example has the resolution of $96 \times 96$ and a binary label indicating the presence of metastatic tissue. The LocalTCT dataset is provided by the cervical cancer pathology screening of local hospitals, which contains about 16,902 WSIs at $20\times$ magnification.  Six categories patches with the resolution of $224\times 224$ are collected and annotated by three professional pathologists, including 18,912 HSILs, 20,599 LSILs, 23,488 ASC-Hs, 21,869 ASC-USs, 2,743 AGCs and 87,592 negative cells.  We divide them into train\&validation\&test datasets, consisting of 110,701, 32,253, and 32,249 patches, respectively. To be adaptive for application scenarios, the test set is sampled from more than 6 hospitals that are not included in the training and validation sets. Moreover, the proposed corruptions are applied to the validation set of these two datasets, generating two corruption benchmark datasets, PatchCamelyon-C and LocalTCT-C. 
	
	\noindent\textbf{DNNs choice.}
	The robustness of ten CNNs and three vision transformers is investigated. Ten CNNs include classical (Alexnet \cite{krizhevsky2012imagenet}, VGG16 \cite{simonyan2014very}, ResNet18 \cite{he2016deep}, ResNet34, ResNet50, and ResNet101), lightweight (MobileNetV2 \cite{sandler2018mobilenetv2} and shuffleNet \cite{zhang2018shufflenet}), and SOTA models (EffecientNetB0 \cite{tan2019efficientnet} and EffecientNetB7). Vision transformers cover ViT \cite{dosovitskiy2020image} and its newest variants, SwinTransformer \cite{liu2021swin} and DeiT \cite{touvron2021training}. Three vision transformers are not applied to the PatchCamelyon due to their unstable training procedure at the resolution of $96 \times 96$.

	\begin{table*}
		\centering
		\caption{Robustness performance on the Patchcamelyon-C and LocalTCT-C. The first and second best results are emphasized by \underline{\textbf{value}} and \textbf{value}, respectively. Results reported by aggregating MAE numbers over 3 different seeds.}
		\begin{tabular*}{\textwidth}{@{\extracolsep{\fill}}l|llll|llll}
			\multicolumn{1}{c}{}&\multicolumn{4}{c}{PatchCamelyon} & \multicolumn{4}{c}{LocalTCT} \\
			\hline
			Method & Error(\%) & CE(\%) & rCE & CEC(\%) & Error(\%) & CE(\%) & rCE & CEC(\%) \\
			\hline
			AlexNet        & 14.94 & 26.87& \underline{\textbf{1.80}}& 53.85&16.06&30.64&\underline{\textbf{1.91}}&43.25 \\
			VGG16          & 10.34 &  \underline{\textbf{23.16}} &  2.24 &  48.56&13.51&30.24&2.24&44.16 \\
			ResNet18       & 11.12 &  24.84 &  2.23 &  44.70&13.23&30.48&2.30&41.41 \\
			ResNet34       & 11.22 &  \textbf{23.60} &  2.10 &  43.47&12.92&28.38&2.20&\textbf{38.89} \\
			ResNet50       & 12.54 &  28.84 &  2.30 &  47.00&12.96&31.61&2.44&41.41 \\
			ResNet101      & 12.00 &  25.15 &  2.09 &  \textbf{43.40}&13.25&29.64&2.24&\underline{\textbf{36.29}} \\
			MobileNetV2    & 9.68 &  26.60 &  2.75 &  44.30&12.81&30.24&2.36&43.61 \\
			ShuffleNet     & 13.53 &  26.15 &  \textbf{1.93} &  53.56&14.53&32.91&2.26&46.98 \\
			EfficientNetb0 & 10.76 &  26.19 &  2.43 &  \underline{\textbf{42.75}}&12.96&30.51&2.35&46.70 \\
			EfficientNetb7 & 10.39 &  24.89 &  2.39 &  43.75&12.56&\underline{\textbf{26.36}}&2.10&44.11 \\
			ViT & -&-&-&-&15.30&30.91&2.02&48.23 \\
			Swin Trans. & -&-&-&-&13.00&28.47&2.19&45.08 \\
			DeiT & -&-&-&-&13.85&\textbf{27.81}&\textbf{2.01}&43.36 \\
			\hline
		\end{tabular*}
		\label{tab1}
	\end{table*}
	
	\subsection{Experimental results}
	\noindent\textbf{Proposed corruptions are close to reality while keeping the diagnostic information in human experts' opinion.} In Section \ref{sec22}, corruption types have been discussed that they are common in reality. Herein, the design of the corruption severity level is also explored to ensure its clinical significance. Otherwise, the diagnostic information remaining in the corrupted images is also crucial to the clinical value of proposed corruptions. 3 images with 45 corruptions each (135 images in total) for two datasets are evaluated by experts. 127 Patchcamelyon and 130 LocalTCT patches are thought likely to appear in the real world. Meanwhile, 118 Patchcamelyon and 127 LocalTCT patches can be correctly recognized by experts.
	
	\noindent\textbf{Image corruptions decrease the prediction accuracy.} The robustness performance of models is shown in Tab. \ref{tab1}. The related metrics on classification performance, Error, CE, and rCE, are first discussed.  rCE values of all models range from 1.8 to 2.8. In other words, the error on corrupted images is 0.8 to 1.8 times larger than that on clean images, indicating the poor robustness of modern DNNs. On both datasets, AlexNet shows the worst Error on all CNNs while it has the best rCE value.  Overall, its CE is comparable to the other CNNs. That is, although  CNNs are constantly improved in the past decade, their performance on corrupted images changes little while causing the incredibly worse robustness. Moreover, three vision transformers have smaller rCE values than most CNNs, illustrating their better robustness towards image corruptions. 
	
	\begin{figure*}[!h]
		\centering
		\subfigure{\includegraphics[width=0.25\columnwidth]{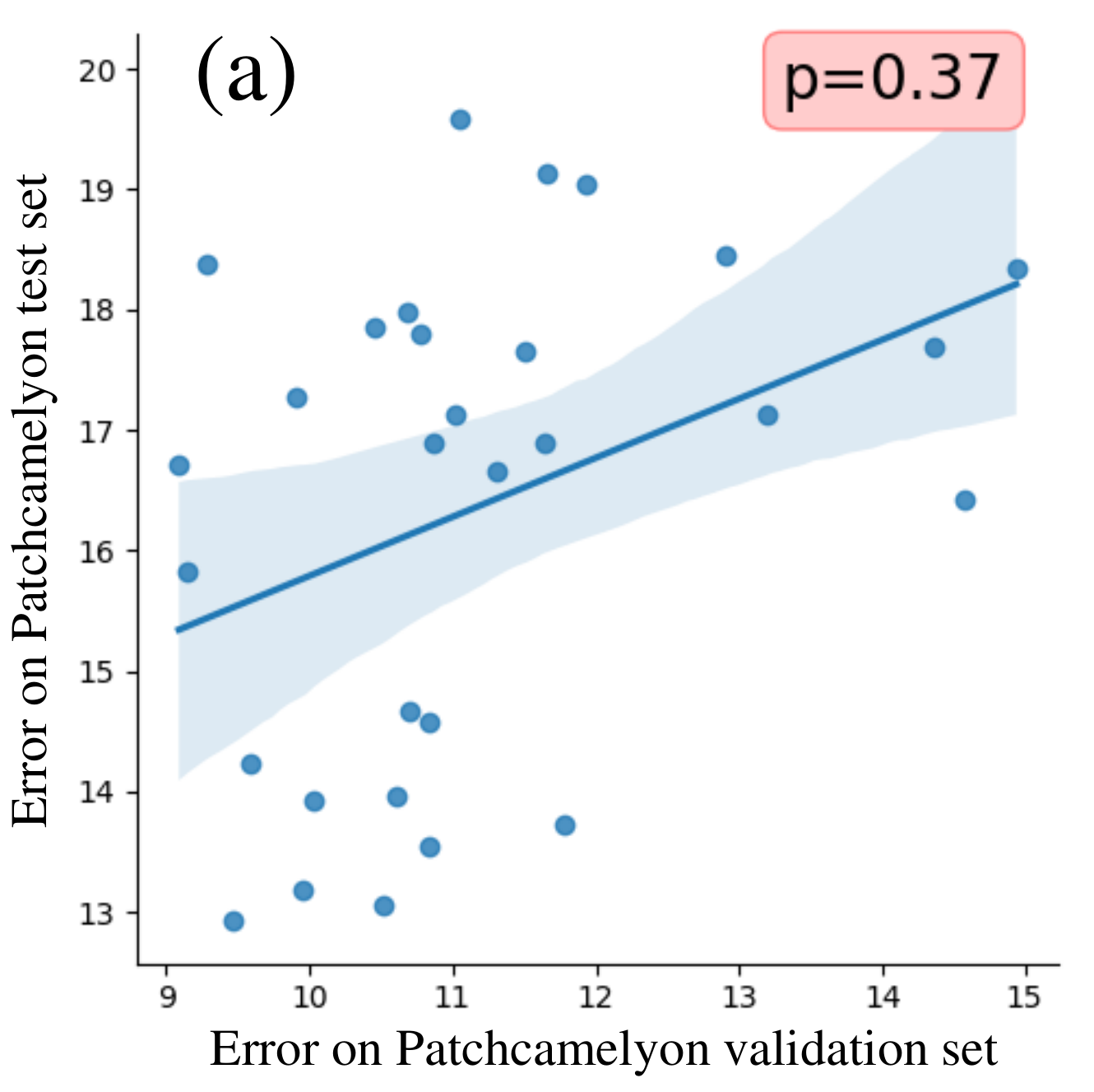}\label{fig11}}\subfigure{\includegraphics[width=0.25\columnwidth]{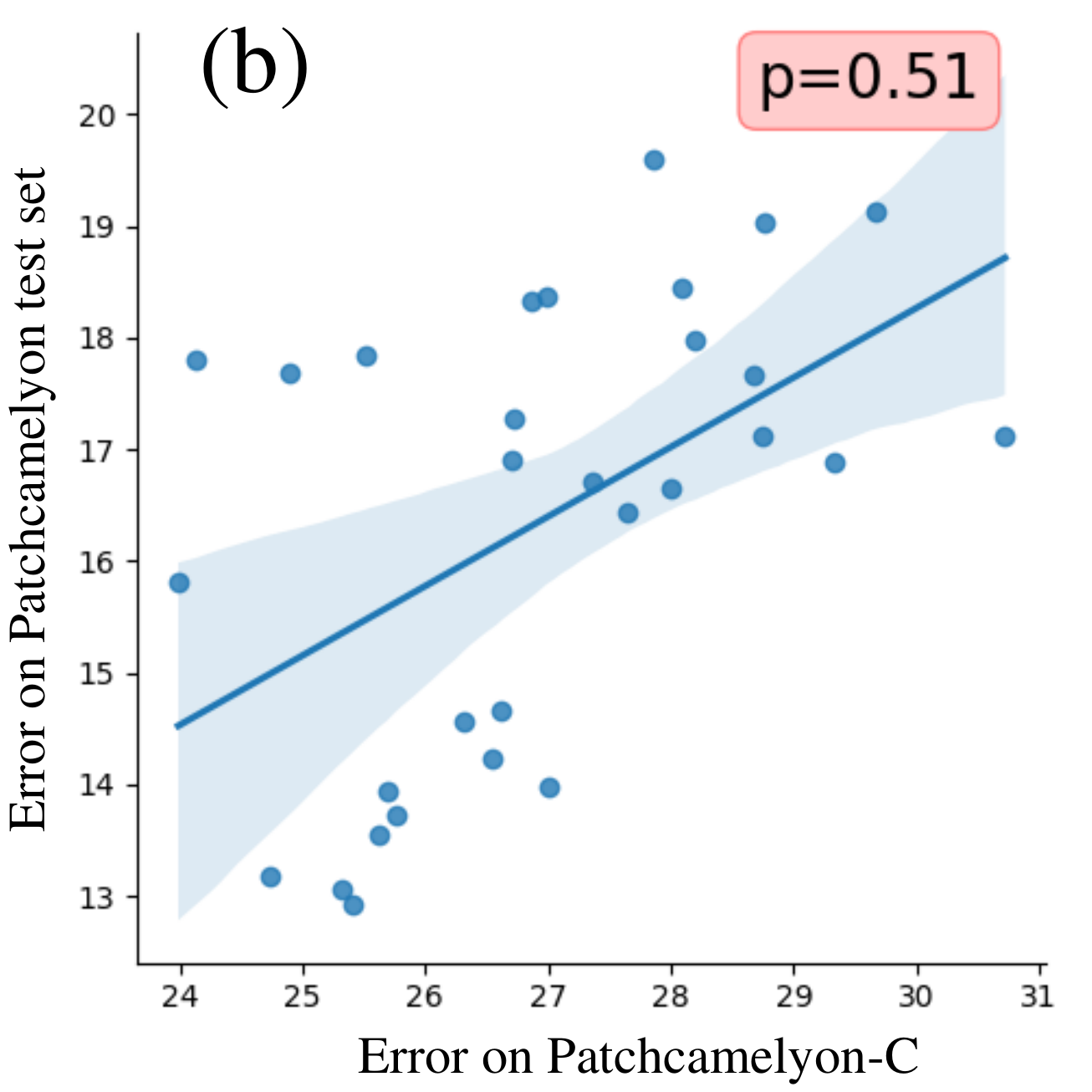}\label{fig12}}\subfigure{\includegraphics[width=0.25\columnwidth]{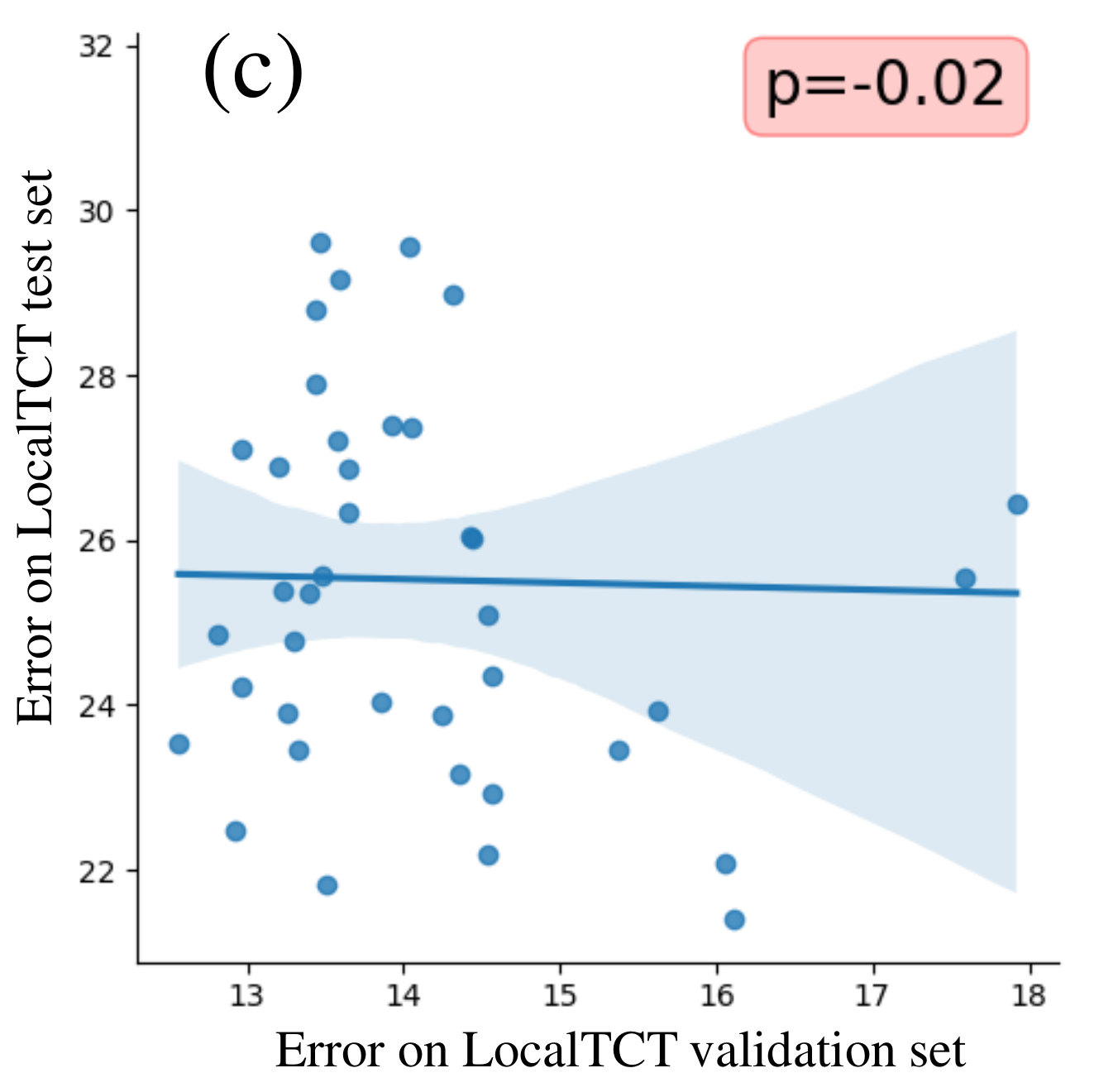}\label{fig13}}\subfigure{\includegraphics[width=0.25\columnwidth]{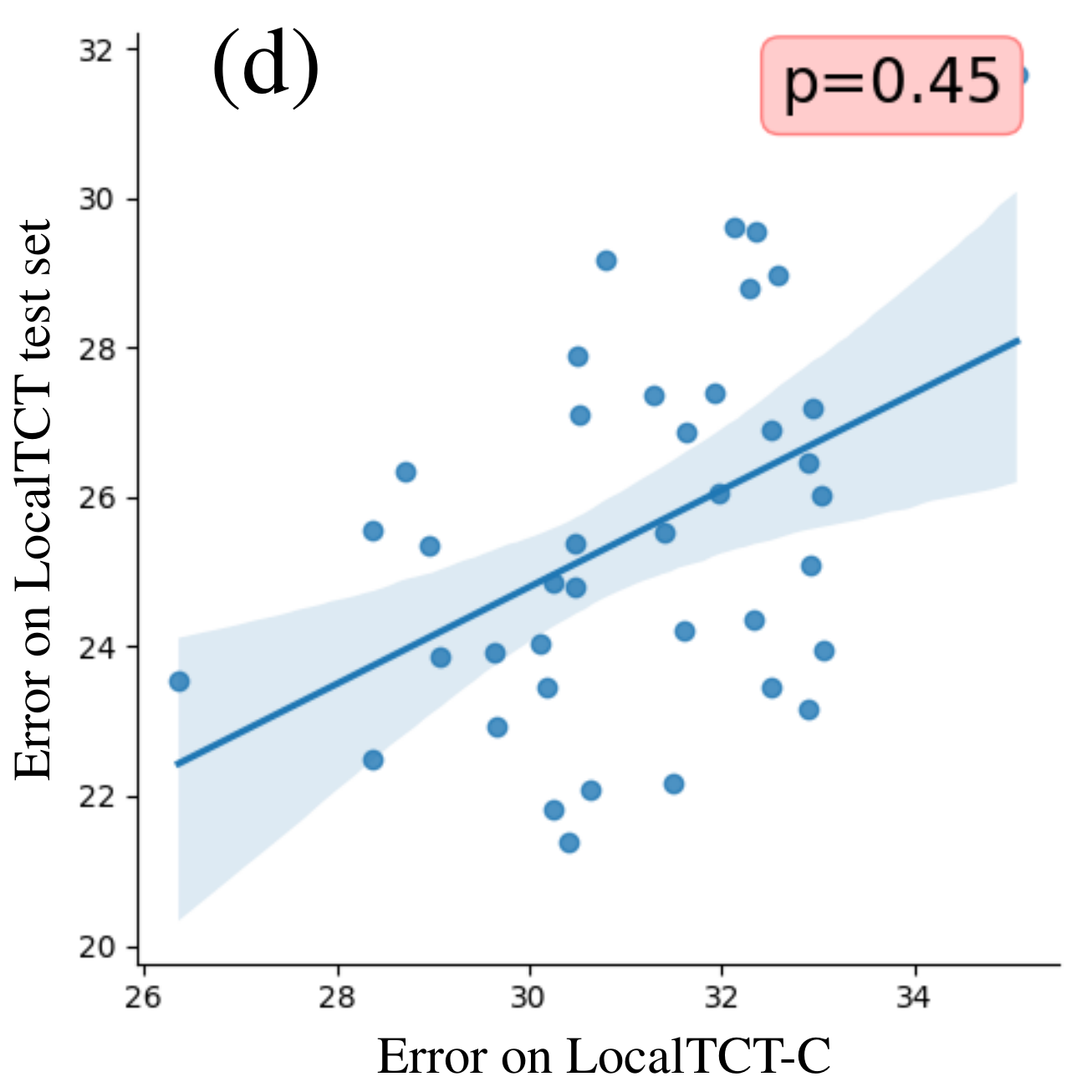}\label{fig14}}
		\caption{Pearson correlation coefficients between the error on Patchcamelyon validation and test sets (a), Patchcamelyon-C and Patchcamelyon test set (b),  LocalTCT validation and test sets (c), and LocalTCT-C and LocalTCT test set (d). Four correlation coefficients are all calculated by 30 points. Results show that the performance on our benchmark is more closely related to that on the test set. 
		}\label{fig3}
	\end{figure*}	
	
	\noindent\textbf{Model confidence is unstable to image corruptions.} We analyze the performance of the CEC and find that CEC values of all models range from 35\% to 55\%, which means nearly half of all confidence values are falsely ranked. This phenomenon manifests the terrible confidence estimation of modern DNNs. Comparing CEC values of CNNs and vision transformers, it is found that the confidence of vision transformers is less reliable than that of most CNNs. Another interesting observation is that the CEC and rCE seem to be in the opposite direction. In other words, if a model has better robustness of the prediction, its robustness of the confidence is worse.

	\noindent\textbf{Corruption types affect model performance.} Comparing the performance on only one type of corruption (shown in Tab. A1 and Tab. A2 of Appendix), more phenomena deserve attention. Firstly, the architecture is robust to some corruptions while vulnerable to others. For example, as shown in Tab. A1, the accuracy of AlexNet changes slightly when encountering the \textit{JPEG} (i.e., error of 20.99\%), whereas \textit{hue} will severely impair its accuracy (i.e., error of 38.50\%). The reason causing this may be that AlexNet is sensitive to the statistical variation while is robust to texture details. Secondly, corruption type has effects on the performance of DNNs. The top half of Tab. A1 shows that most architectures' CE values below 20\% on the corruption of \textit{mark} while having CE values above 30\% on the corruption of \textit{defocus blur}. Thirdly, the performance of models, especially the CEC, varies widely on the same corruption. As shown in the top half of Tab. A2, the best CEC is 48.31\% and the worst one is 72.54\% when the \textit{hue} varies.

	\noindent\textbf{Early-stopping helps improve robustness towards corruptions.} The early-stopping is a regularization technique used to avoid overfitting when the validation error does not drop. We find that overfitting also harms the corruption robustness of models. As shown in Fig. A1 of Appendix, although the validation error has stabilized after the 4,000 iterations,  CE values rise one to two points compared to the best results at the 4,000 iterations. Thus, early-stopping is also helpful for improving models' robustness with respect to image corruption.
	
	\noindent\textbf{Robustness against corruptions is related to the generalization ability of models.} Fig. \ref{fig3} (a)-(d) show the correlation between the error on the validation set or our benchmark, and the error on the test set. On the Patchcamelyon dataset, the proposed benchmark has a higher Pearson correlation coefficient than the clean validation set (0.51 v.s., 0.37). On the LocalTCT dataset, the correlation coefficient between the validation and test errors is $-0.02$, which means no correlation or even slightly negative correlation between them. In other words, the validation set loses its ability to choose the model with good generalization. In comparison,  the correlation coefficient between the error on LocalTCT-C and LocalTCT test set rises to 0.45. Hence, our benchmark can be used to estimate which model has better generalization ability.
	
	\section{Conclusion}

	In this paper, we present a new benchmark to evaluate the robustness of current DNNs with respect to real-world pathology image corruptions. The value of the proposed benchmark is explored from two aspects. Firstly, it is crucial to guarantee the robustness of the model for pathology image analysis. However, there is still considerable room for improving the models' robustness against image corruption. Hence, we encourage more efforts to be devoted to improving the models' robustness using our benchmark. Secondly, experimental results show that the proposed benchmark is more reliable than the validation set in accurately evaluating the generalization behavior of models, which is useful for practitioners to design the right model for their task at hand.
	
	\noindent\textbf{Limitations and Future Work.} Our goal is to investigate the performance of DNNs on corrupted images, but the role of input augmentation is neglected in this paper. In fact, augmentation is verified to be able to improve the robustness towards corruption in the field of digital pathology \cite{yamashita2021learning}. Moreover, strong augmentations may be the recipe of the big architectures in improving robustness \cite{bai2021transformers}. Next, we will analyze the impact of augmentation on corruption robustness.
	
	\noindent\textbf{Acknowledgements.} This work was funded by China Postdoctoral Science Foundation (2021M702922).

	\bibliographystyle{splncs04}
	\bibliography{miccai2022}
\end{document}